\documentclass[runningheads]{llncs}

\usepackage[T1]{fontenc}

\usepackage{graphicx,verbatim}

\usepackage{booktabs}
\usepackage{multirow}

\usepackage{amsmath,amssymb}

\usepackage{url}
\usepackage[misc]{ifsym}
\newcommand{\first}[1]{\textbf{\underline{#1}}}
\newcommand{\second}[1]{\textbf{#1}}
\newcommand{\third}[1]{\textbf{\textit{#1}}}

\begin{document}

\title{Understanding Synergistic Interactions among Pathology Foundation Models via Adaptive Fusion}
\titlerunning{Synergistic Interactions among Pathology FMs via AdaFusion}

\author{
Yuxiang Xiao\inst{1}\thanks{Yuxiang Xiao and Yang Hu contribute equally to this work.\\
Corresponding email: \email{superhy199148@hotmail.com} and
\email{yangkx@scut.edu.cn}} \and
Yang Hu\inst{2,3,4\star}\textsuperscript{(\Letter)} \and
Bin Li\inst{4} \and
Tianyang Zhang\inst{4} \and
Zexi Li\inst{6} \and
Huazhu Fu\inst{7} \and
Jens Rittscher\inst{4,5} \and
Kaixiang Yang\inst{1}\textsuperscript{(\Letter)}
}

\authorrunning{Yuxiang Xiao, Yang Hu et al.}

\institute{
School of Computer Science and Engineering, South China University of Technology, China
\and
School of Computing and Mathematical Sciences, University of Leicester, UK
\and
Leicester Cancer Research Centre, University of Leicester, UK
\and
Department of Engineering Science, University of Oxford, UK
\and
Nuffield Department of Medicine, University of Oxford, UK
\and
St Catherine's CollegeUn, University of Oxford, UK
\and
A*STAR, Singapore
}

\maketitle

\begin{abstract}
Pathology foundation models (PFMs) provide strong tile-level representations via self-supervised pre-training on large-scale pathology images. Yet, PFMs are developed under diverse and often opaque data, architecture, and objective choices, inducing latent representational biases that limit robustness and obscure what each model specialises in.
We present \textbf{AdaFusion}, a lightweight adaptive fusion framework that integrates complementary signals from multiple frozen PFMs through (1) low-dimensional feature compression and (2) a sample-conditioned gating module that reweights model-wise (and optionally channel-wise) contributions.
Beyond improving predictive accuracy, AdaFusion provides \emph{contribution-driven interpretation} that offers evidence consistent with model-specific preferences and synergistic interactions across tissue phenotypes.
We evaluate AdaFusion on three public benchmarks spanning treatment response prediction, prostate cancer grading, and spatial gene expression inference. AdaFusion consistently outperforms individual PFMs and other fusion baselines, while providing interpretable tissue visualisation which aligns model preferences with morphological patterns.
Code is available at: \url{https://github.com/xyx-98/PathoOracle}.
\end{abstract}

\section{Introduction}
\label{sec:intro}

High-resolution digital pathology images provide rich tissue context and cellular detail essential for computational modelling in cancer diagnosis, grading, and prognosis~\cite{lee2025benchmarking}. In weakly supervised slide-level settings, multiple instance learning (MIL) decomposes gigapixel whole slide images (WSIs) into tiles and aggregates tile embeddings to support downstream tasks such as subtyping, grading, and prognosis~\cite{ilse2018attention,lu2021data,chen2024towards,hu2025self}. The quality of tile-level features is a critical determinant of downstream performance. For a long time, tile-level embeddings were typically extracted using encoders pretrained on conventional natural-image benchmarks (e.g., ImageNet~\cite{deng2009imagenet}).

Recently, the rise of self-supervised learning (SSL)-based pathology foundation models (PFMs) has revolutionised the extraction of tile representations~\cite{chen2024towards,xu2024whole,vorontsov2024foundation}. These models, pretrained on massive pathological image cohorts, provide high-quality in-domain tile embeddings that can easily integrate with multimodal information such as textual clinical reports or genomic profiles~\cite{lu2024visual,xu2025multimodal}. The widespread deployment of PFMs has ushered in a new era in computational pathology, shifting the focus from task-specific learning towards generalisable, feature-centric pipelines~\cite{wang2024pathology}.

Despite these advancements, existing PFMs are often developed under diverse and opaque training conditions, introducing potential biases that stem from limited data diversity and inaccessible pretraining sources~\cite{chen2024towards,lu2024visual,xu2024whole}. Due to patient privacy regulations, most of the private datasets used to train PFMs are not publicly available, making it difficult to assess their quality, discrepancies, or underlying distributions. While some PFMs claim to be trained on multi-centre datasets, key demographic details, such as sampling geography, ethnicity, and gender, are frequently unknown~\cite{vorontsov2024foundation,wang2024pathology}. Even among those with disclosed training sources, the data are typically collected from one or two regional hospitals, and often display pronounced class imbalance across cancer types, following long-tailed distributions~\cite{chen2024towards,vorontsov2024foundation}.

Beyond data-induced biases, the structural design and training algorithms of PFMs introduce additional representational biases. The widely used Vision Transformer (ViT) processes image patches as independent tokens~\cite{chen2024towards,zimmermann2024virchow2}, promoting local feature learning but limiting long-range spatial modelling. In contrast, emerging dilated ViT architectures~\cite{xu2024whole} emphasise global context and tissue architecture, often at the cost of cytological detail. Meanwhile, the choice of self-supervised objective further shapes representations: contrastive approaches like DINOv2 instil discriminative priors, while masked autoencoding (MAE) encourages holistic, context-aware features. Visual--language models introduce semantic priors by aligning features with language~\cite{lu2024visual}, but may overlook morphology beyond linguistic description. The choice of structure and pre-training methods also yields PFMs with specialised yet limited perspectives.

\begin{figure}[t]
\centering
\includegraphics[width=\textwidth]{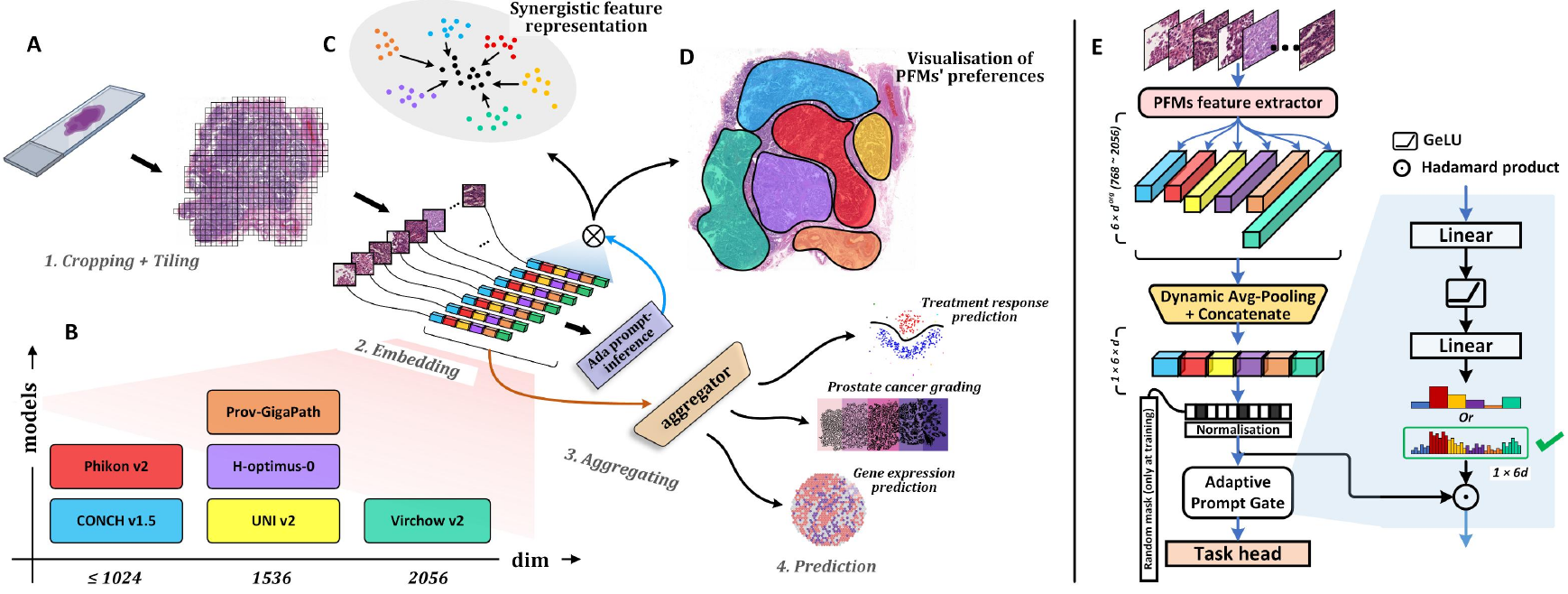}
\caption{\textbf{AdaFusion overview and adaptive fusion gate.}
(A) Standard WSI workflow: tiling, tile embedding, aggregation, and downstream prediction.
(B) PFMs output heterogeneous embedding dimensions.
(C) Adaptive fusion yields a synergistic representation by integrating complementary PFM signals.
(D) Contribution visualisation highlights regions where different PFMs dominate, consistent with model-specific preferences.
(E) Architecture of the adaptive fusion gate: pooled per-PFM features are concatenated and reweighted by a lightweight gating module before the task head.}
\label{fig:overview}
\end{figure}

Such biases are silently propagated through pretrained feature spaces, compromising the robustness and fairness of downstream analyses~\cite{du2025ethics}. Given the proliferation of diverse PFMs, integrating their complementary strengths presents an intuitive yet underexplored direction. In this work, we focus on \emph{understanding and leveraging complementary information among PFMs}, offering a synergy-aware view of multi-PFM interactions: rather than selecting a single “best” PFM, we learn a lightweight adaptor that coordinates multiple frozen PFMs and provides model-wise contribution scores that are informative about tissue phenotypes. \textbf{We summarise the contributions of this study as:}
(1) We develop \textbf{AdaFusion}, an embarrassingly simple but efficient adaptive fusion framework that compresses heterogeneous PFM embeddings and applies sample-conditioned gating for multi-PFM integration (Fig.~\ref{fig:overview}).
(2) We introduce \textbf{contribution-driven interpretability} that provides evidence consistent with model-specific preferences across tissue regions.
(3) We validate AdaFusion on diverse benchmarks (classification and regression), demonstrating consistent performance gains and clinically meaningful interpretability.

\section{Related Work}
\label{sec:rw}

Early computational pathology often relied on natural-image pretraining (e.g., ImageNet)~\cite{lu2021data}. With the rise of self-supervised learning (SSL), pathology-specific encoders pretrained on unlabelled WSI data improved transferability and downstream performance~\cite{dehaene2020self}. 

More recently, pathology foundation models (PFMs) scale SSL pretraining to larger and more diverse WSI corpora, producing general-purpose pan-cancer histomorphological representations. Representative examples include CTransPath~\cite{wang2022transformer}, UNI~\cite{chen2024towards}, Phikon-v2~\cite{filiot2024phikon}, and Virchow~\cite{zimmermann2024virchow2} (DINOv2-pretrained~\cite{oquab2024dinov2}). Larger PFMs exceed a billion parameters (e.g., H-optimus-0~\cite{hoptimus0}, Prov-GigaPath~\cite{xu2024whole}), and some incorporate multimodal signals to enhance biological relevance (e.g., CONCH~\cite{lu2024visual}, mSTAR~\cite{xu2025multimodal}). 

Despite their scale, PFMs may still inherit batch effects across centres and acquisition pipelines, limiting robustness and out-of-distribution generalisation~\cite{vaidya2024demographic}. Prior studies mitigate these issues via downstream fine-tuning~\cite{lee2025benchmarking}, text-based supervision~\cite{guo2025focus}, robustness strategies such as random PFM sampling during feature extraction~\cite{lenz2025unsupervised}, and combining a small subset of PFM features~\cite{gao2025features}. However, existing approaches provide limited interpretability of \emph{which} PFM contributes \emph{where}, and rarely examine synergistic roles of PFMs across tissue phenotypes.

\section{Method}
\label{sec:method}


AdaFusion integrates multiple frozen PFMs via a lightweight adaptive fusion gate (Fig.~\ref{fig:overview}E), targeting: (1) \textbf{efficiency} through low-dimensional compression of heterogeneous embeddings; (2) \textbf{adaptivity} through sample-conditioned reweighting of model contributions; and (3) \textbf{interpretability} via model-wise contribution scores. Adafusion has these \textbf{differences to prior fusion schemes:}
Unlike naive concatenation, token-level self-attention, or top-$k$ MoE selection, AdaFusion applies a compact sample-conditioned gate (model-wise or model\,$\times$\,channel) with dense soft weights to reweight frozen PFMs.

\subsection{Problem Setting}
We consider both slide-level weakly supervised learning (MIL) and tile-level prediction.
For WSI-based MIL, we segment foreground tissue and divide each WSI into non-overlapping tiles of size $224 \times 224$ pixels. Following the MIL paradigm, the $n$-th WSI is represented as a bag
$\mathcal{X}_n = \{\mathbf{x}_{n,i}\}_{i=1}^{M_n}$,
where $\mathbf{x}_{n,i}$ is the $i$-th tile and $M_n$ is the number of tiles.

\subsection{Multi-PFM Feature Extraction and Compression}
Given a tile $x \in \mathbb{R}^{H \times W \times C}$, we extract features from $N$ frozen PFMs $\{\mathcal{F}_1,\dots,\mathcal{F}_N\}$. Each PFM $\mathcal{F}_i$ maps the input to an embedding $\mathbf{f}_i \in \mathbb{R}^{d_i}$, via $\mathbf{f}_i = \mathcal{F}_i(x; \theta_{\mathcal{F}_i})$, where $\theta_{\mathcal{F}_i}$ is fixed and $d_i$ varies across PFMs.

To enable efficient fusion across heterogeneous dimensions, we compress each $\mathbf{f}_i$ to a unified dimension $d$ (default $d{=}64$) and produce a set of standardised per-model embeddings $\{\mathbf{e}_1,\dots,\mathbf{e}_N\}$ for each tile, via mean pooling: 
\begin{equation}
\mathbf{e}_i = \mathrm{Pool}(\mathbf{f}_i) \in \mathbb{R}^{d}.
\end{equation}

\subsection{Adaptive Fusion Gate \& Downstream Prediction}
\textbf{Adaptive Fusion Gate.} 
We concatenate the per-model embeddings into a compound representation as $\mathbf{E} = [\mathbf{e}_1;\mathbf{e}_2;\dots;\mathbf{e}_N] \in \mathbb{R}^{N \times d}.$
To improve robustness, we apply random masking during training:
\begin{equation}
\mathbf{E}_{\text{mask}} = \mathbf{E} \odot \mathbf{M},
\end{equation}
where $\mathbf{M}\in\{0,1\}^{N\times d}$ keeps entries with probability $1-\rho$ (default $\rho{=}0.2$).

We learn a lightweight gating module $\mathcal{G}(\cdot;\theta_g)$ that outputs a sample-conditioned attention value tensor:
\begin{equation}
\mathbf{W}_{\text{gate}} = \mathcal{G}(\mathbf{E}_{\text{mask}};\theta_g),
\end{equation}
and two variants are considered:
(1) \textbf{AdaFusion-Coarse:} (AdaFu-C) $\mathbf{W}'_{\text{gate}} \in \mathbb{R}^{N\times 1}$ produces model-wise weights, broadcast to $\mathbb{R}^{N\times d}$.
(2) \textbf{AdaFusion-Fine:} (AdaFu-F) $\mathbf{W}_{\text{gate}} \in \mathbb{R}^{N\times d}$ produces channel-wise weights for finer control.

The fused representation is obtained by element-wise Hadamard product for reweighting and passing to the downstream head:
\begin{equation}
\mathbf{Z} = Flatten(\mathbf{E} \odot \mathbf{W}_{\text{gate}}), \ Z \in \mathbb{R}^{Nd}.
\end{equation}

\noindent\textbf{Downstream Prediction.} 
For tile-level tasks, we use a linear head $\mathcal{H}(\cdot)$ to obtain predictions as
$\hat{y} = \mathcal{H}(\mathbf{Z};\theta_h).$
In slide-level, each tile is fused into $\mathbf{Z}_{n,i}$ and aggregated via an attention-based MIL module~\cite{ilse2018attention}, followed by a linear classifier.

\subsection{Contribution-Driven Interpretability}
The learned gate of AdaFusion exposes how PFMs contribute under different tissue phenotypes. We decompose $\mathbf{W}_{\text{gate}}$ into per-model blocks $\{\mathbf{w}_1,\dots,\mathbf{w}_N\}$ with $\mathbf{w}_i\in\mathbb{R}^{d}$.
We compute a scalar contribution score for model $\mathcal{F}_i$:
\begin{equation}
S_i = \frac{1}{d}\sum_{j=1}^{d} (\mathbf{w}_i)_j.
\end{equation}

For a WSI with $M$ tiles, AdaFusion yields a slide-level contribution tensor $\mathbf{A}\in\mathbb{R}^{M\times N}$, where $\mathbf{A}_{m,i}=S_i(\mathbf{x}_m)$ denotes the (scalar) contribution score of PFM $\mathcal{F}_i$ for tile $\mathbf{x}_m$. We then assign each tile $\mathbf{x}_m$ the index of the PFM with maximal $\mathbf{A}_{m,i}$ and render spatial contribution maps (Fig.~\ref{fig:results_vis}), enabling phenotype-linked analysis and providing evidence consistent with complementary model preferences across PFMs.


\section{Results}
\label{sec:results}

\subsection{Datasets \& Experimental Setting}
We evaluate our method on three public benchmarks spanning binary classification, multi-class classification, and regression tasks: (1) \textbf{ATEC23 (ATEC2023-Challenge)}~\cite{wang2025atec23}: an ovarian cancer dataset for predicting response to bevacizumab (effective vs.\ ineffective). WSIs are sourced from multiple centres. We use 5-fold cross-validation. (2) \textbf{PANDA (Prostate cAncer graDe Assessment)}~\cite{bulten2022artificial}: a large-scale prostate cancer dataset with 10,616 WSIs annotated by ISUP grade (0--5). We perform 5-fold cross-validation on the development set. (3) \textbf{HEST-Benchmark (Histology-based Gene Expression Signature Test)}~\cite{jaume2024hest}: a regression benchmark of nine tasks predicting expression levels of the top 50 spatially variable genes across cancer types. We follow the official training--testing splits.

\subsection{Implementation Details}
\noindent\textbf{Feature Extraction.} We use six PFMs (CONCH v1.5~\cite{lu2024visual}, Phikon v2~\cite{filiot2024phikon}, UNI v2~\cite{chen2024towards}, H-optimus-0~\cite{hoptimus0}, Prov-GigaPath~\cite{xu2024whole} (p-GigaPath), and Virchow v2~\cite{zimmermann2024virchow2}). The parameter-free pooling is used to produce a unified, compressed tile embedding, which avoids dataset-specific feature selection and preserves the pretrained distribution.

\noindent\textbf{Task Head.} For classification tasks (ATEC23 and PANDA), tile features are aggregated into slide-level representations via an attention-based MIL (ABMIL) model~\cite{ilse2018attention}, followed by a linear classifier. For regression (HEST-Benchmark), fused features are passed to a linear head to predict gene expressions.

\noindent\textbf{Training Setup.} All trainable modules (adaptive fusion gate and task head) are optimised using Adam with a learning rate of $2\mathrm{e}{-4}$ and weight decay of $1\mathrm{e}{-5}$. Classification uses cross-entropy loss (50 epochs), and regression uses mean squared error (20 epochs). All models are trained on a single NVIDIA 4090 GPU.

\noindent\textbf{Evaluation Metrics.} We report Accuracy (ACC) and Area Under the ROC Curve (AUC) for classification tasks, and Pearson Correlation Coefficient (PCC) for regression. For cross-validation, mean scores are reported.

\noindent\textbf{Baselines.} We compare AdaFusion with every single PFM feature and two fusion baselines: (1) \textbf{Self-Attn}~\cite{gao2025features}, which applies standard self-attention to concatenated features; and (2) \textbf{Top-3 MoE}, a mixture-of-experts model with a gating mechanism that selects the top three features for concatenation and remapping via a linear layer.

\begin{table}[t]
\caption{Performance on classification benchmarks (ATEC23 and PANDA). Individual PFMs are evaluated with original dimensions (ori-d) and pooled dimensions (512/64). Fusion baselines and AdaFusion use pooled features. First/second/third best results are marked by bold-underline / bold / bold-italic.}
\fontsize{8pt}{9.6pt}\selectfont
\centering
\setlength{\tabcolsep}{0.5mm}        
\setlength{\belowrulesep}{0pt}        
\setlength{\aboverulesep}{0pt}

\begin{tabular}{@{}l|ccc|ccc|ccc|ccc@{}}
\toprule
\multirow{2}{*}{\textbf{Model}} &
\multicolumn{3}{c|}{\textbf{ATEC23 ACC}} &
\multicolumn{3}{c|}{\textbf{ATEC23 AUC}} &
\multicolumn{3}{c|}{\textbf{PANDA ACC}} &
\multicolumn{3}{c}{\textbf{PANDA AUC}} \\
& ori-d & 512 & 64 & ori-d & 512 & 64 & ori-d & 512 & 64 & ori-d & 512 & 64 \\
\midrule
CONCH v15 & 0.779 & 0.779 & 0.724 & 0.795 & 0.798 & 0.745 & 0.673 & 0.674 & 0.641 & 0.909 & 0.908 & 0.885 \\
Phikon v2 & 0.828 & 0.810 & 0.779 & 0.861 & 0.873 & 0.817 & 0.721 & 0.713 & 0.623 & 0.928 & 0.923 & 0.872 \\
UNI v2 & 0.831 & 0.824 & 0.724 & 0.889 & 0.880 & 0.779 & 0.739 & 0.717 & 0.643 & 0.936 & 0.922 & 0.888 \\
H-optimus-0 & 0.855 & 0.803 & 0.741 & 0.913 & 0.851 & 0.783 & 0.753 & 0.728 & 0.681 & 0.942 & 0.930 & 0.908 \\
p-GigaPath & 0.841 & 0.797 & 0.772 & 0.868 & 0.849 & 0.819 & 0.742 & 0.714 & 0.666 & 0.936 & 0.924 & 0.897 \\
Virchow2 & 0.852 & 0.841 & 0.721 & 0.885 & \second{0.905} & 0.776 & 0.740 & 0.722 & 0.657 & 0.938 & 0.928 & 0.892 \\
\midrule
Self-Attn & - & \second{0.855} & \third{0.838} & - & 0.890 & \third{0.886} & - & 0.727 & \third{0.754} & - & 0.926 & \third{0.942} \\
Top-3 MoE & - & \third{0.852} & 0.828 & - & 0.873 & 0.880 & - & \third{0.731} & 0.718 & - & \third{0.934} & 0.927 \\
AdaFu-C & - & 0.841 & \first{0.862} & - & \third{0.896} & \first{0.901} & - & \second{0.793} & \second{0.792} & - & \second{0.952} & \second{0.951} \\
AdaFu-F & - & \first{0.869} & \first{0.862} & - & \first{0.910} & \second{0.900} & - & \first{0.821} & \first{0.815} & - & \first{0.958} & \first{0.955} \\
\bottomrule
\end{tabular}
\label{tab:classification_results}
\end{table}

\begin{table}[t]
\caption{HEST-Benchmark regression results (PCC). PFMs are evaluated at their original dimensions, while fusion methods use a $6 \times 64$ representation. First/second/third best results are marked by bold-underline / bold / bold-italic.}
\fontsize{8pt}{9.6pt}\selectfont
\centering
\setlength{\tabcolsep}{0.8mm}         
\setlength{\belowrulesep}{0pt}
\setlength{\aboverulesep}{0pt}

\begin{tabular}{@{}l|ccccccccc|c@{}}
\toprule
\textbf{Model} & \textbf{ID} & \textbf{PR} & \textbf{PA} & \textbf{SK} & \textbf{CO} & \textbf{RE} & \textbf{RC} & \textbf{LU} & \textbf{LY} & \textbf{Avg} \\
\midrule
CONCH v15 & 0.504 & \third{0.373} & 0.391 & 0.460 & \second{0.264} & 0.157 & 0.188 & 0.501 & \third{0.257} & 0.344 \\
Phikon v2 & 0.536 & 0.303 & 0.410 & 0.495 & 0.221 & 0.146 & \first{0.250} & 0.470 & 0.224 & 0.339 \\
UNI v2 & \third{0.579} & 0.369 & 0.420 & 0.583 & 0.219 & 0.181 & 0.233 & 0.450 & 0.224 & 0.362 \\
H-optimus-0 & \second{0.586} & 0.327 & 0.434 & \first{0.609} & \third{0.257} & 0.192 & \second{0.239} & 0.495 & 0.227 & \third{0.374} \\
p-GigaPath & 0.537 & 0.346 & 0.398 & 0.524 & 0.227 & 0.155 & 0.208 & 0.460 & 0.211 & 0.341 \\
Virchow v2 & \first{0.589} & 0.351 & \first{0.457} & \third{0.599} & 0.242 & \third{0.207} & \third{0.237} & \second{0.559} & 0.242 & \second{0.387} \\
\midrule
Self-Attn & 0.571 & 0.340 & \second{0.454} & 0.596 & 0.214 & 0.203 & 0.225 & \first{0.562} & 0.239 & 0.378 \\
Top-3 MoE & 0.450 & 0.335 & 0.288 & 0.284 & 0.220 & 0.098 & 0.190 & 0.379 & 0.240 & 0.276 \\
AdaFu-C & 0.557 & \second{0.394} & 0.416 & 0.523 & 0.235 & \second{0.227} & 0.203 & 0.482 & \second{0.273} & 0.368 \\
AdaFu-F & 0.565 & \first{0.407} & \third{0.442} & \second{0.600} & \first{0.279} & \first{0.240} & 0.236 & \third{0.542} & \first{0.277} & \first{0.399} \\
\bottomrule
\end{tabular}
\label{tab:hest_detailed_results}
\end{table}

\begin{table}[t]
\caption{Ablation study. We compare AdaFusion-Fine against direct concatenation (Ensemble) and concatenation with random masking. Experiments are run with PFM features pooled to different pre-concatenation dimensions ($d$).}
\fontsize{8pt}{9.6pt}\selectfont
\centering

\setlength{\aboverulesep}{0pt}
\setlength{\belowrulesep}{0pt}

\setlength{\tabcolsep}{1mm}

\begin{tabular}{@{}l|cc|cc|c@{}}
\toprule
\multicolumn{1}{c|}{\multirow{2}{*}{\textbf{Setting}}} &
\multicolumn{2}{c|}{\textbf{ATEC23}} &
\multicolumn{2}{c|}{\textbf{PANDA}} &
\textbf{HEST} \\
\multicolumn{1}{c|}{} & ACC & AUC & ACC & AUC & PCC \\
\midrule
Ensemble (d=64) & 0.807 & 0.864 & 0.720 & 0.928 & 0.321 \\
Ensemble (d=64, w Mask) & 0.841 & 0.873 & 0.743 & 0.938 & 0.336 \\
AdaFusion-Fine (d=64, full) & \textbf{0.862} & \textbf{0.900} & \textbf{0.815} & \textbf{0.955} & \textbf{0.399} \\
\midrule
Ensemble (d=256) & 0.855 & 0.895 & 0.752 & 0.942 & 0.379 \\
Ensemble (d=256, w Mask) & 0.845 & 0.885 & 0.766 & 0.945 & 0.393 \\
AdaFusion-Fine (d=256, full) & \textbf{0.863} & \textbf{0.909} & \textbf{0.817} & \textbf{0.956} & \textbf{0.423} \\
\midrule
Ensemble (d=512) & 0.856 & 0.908 & 0.760 & 0.945 & 0.395 \\
Ensemble (d=512, w Mask) & 0.855 & 0.909 & 0.777 & 0.950 & 0.400 \\
AdaFusion-Fine (d=512, full) & \textbf{0.869} & \textbf{0.910} & \textbf{0.821} & \textbf{0.958} & \textbf{0.414} \\
\bottomrule
\end{tabular}
\label{tab:ablation_method}
\end{table}

\begin{figure}[t]
\centering
\includegraphics[width=\textwidth]{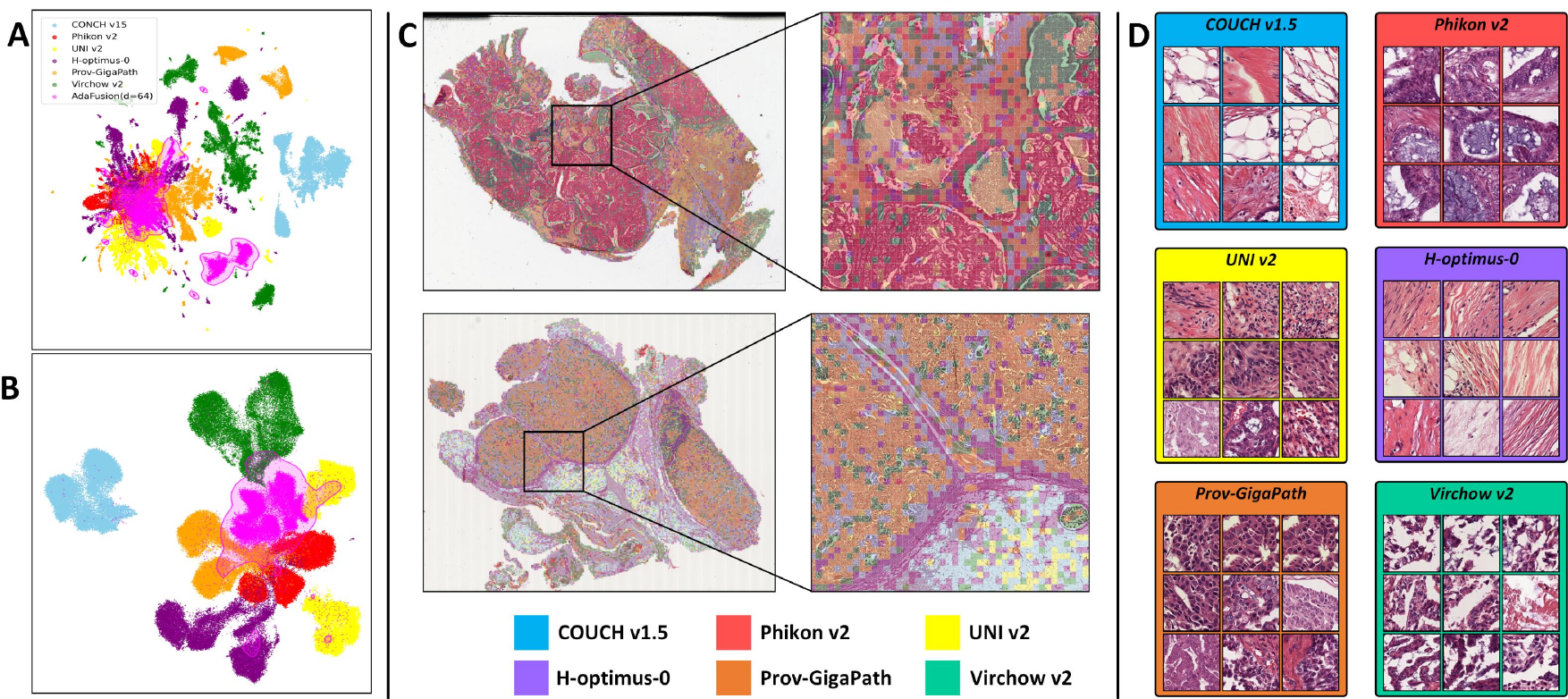}
\caption{\textbf{Synergistic interactions revealed by AdaFusion.}
(Left, A and B) UMAP visualisation of tile features (top: ATEC23; bottom: PANDA) comparing individual PFMs and AdaFusion, showing that the fused representation occupies an integrative position across model-specific feature spaces.
(Right, C and D) Contribution-driven interpretability: spatial maps assign each region to the PFM with the highest contribution score, and exemplar tiles illustrate model-specific morphological preferences. Together, these results are consistent with complementary PFM preferences coordinated by AdaFusion.}
\label{fig:results_vis}
\end{figure}

\subsection{Experimental Results}
We present the comprehensive results of our evaluation in Table~\ref{tab:classification_results} and Table~\ref{tab:hest_detailed_results}. Experiments spanning classification and regression tasks consistently show that AdaFusion outperforms individual state-of-the-art PFMs, highlighting the robustness and generalisability of fusing diverse feature representations.

\noindent\textbf{Results on Classification Tasks.}
Table~\ref{tab:classification_results} benchmarks AdaFusion against individual PFMs and established fusion baselines on ATEC23 and PANDA. On ATEC23, AdaFusion outperforms all PFMs and fusion baselines and remains robust even when feature dimensions are aggressively reduced. The compact $6\times 64$ variants achieve top-tier accuracy, matching higher-dimensional counterparts, indicating that the fusion gate preserves essential information in a highly efficient representation.
The PANDA multi-class grading task further highlights the advantage of adaptive fusion: AdaFusion achieves higher performance under highly compressed $6\times 64$ features, significantly outperforming comparison methods in terms of ACC and AUC. This supports our hypothesis that complementary knowledge across PFMs can be coordinated effectively.

\noindent\textbf{Results on Regression Task.}
On HEST-Benchmark (Table~\ref{tab:hest_detailed_results}), AdaFusion achieves the highest average PCC, outperforming all individual models and fusion baselines. The superiority is driven by consistently strong performance across sub-tasks, suggesting that adaptive reweighting helps exploit complementary inductive biases across PFMs.

\noindent\textbf{Ablation Studies.}
To validate our design choices, we conduct ablations focusing on the contribution of the adaptive gate versus naive concatenation. As shown in Table~\ref{tab:ablation_method}, AdaFusion-Fine consistently outperforms both concatenation baselines across all datasets and dimensionalities, indicating that sample-conditioned reweighting is critical for coordinating complementary model signals. Random masking provides a mild regularisation benefit but remains inferior to learned gating, supporting the necessity of adaptive fusion for robust performance.

\subsection{Visualisation and Interpretability}
\noindent\textbf{Visualisation of Feature Fusion.}
We visualise UMAP projections of tile-level features for ATEC23 and PANDA (Fig.~\ref{fig:results_vis}, A and B). AdaFusion features occupy a central, integrative position rather than forming a separate cluster, indicating that the fused representation effectively integrates complementary information instead of creating another isolated embedding space.

\noindent\textbf{Visualisation of Model Contributions.}
We visualise contribution scores derived from the adaptive gate (Fig.~\ref{fig:results_vis}, C and D). Colour-coded regions align with distinct histomorphological phenotypes, suggesting that different PFMs exhibit preferences for specific tissue structures (e.g., tumour, stroma, glandular regions). Exemplar tiles illustrate these preferences.

\section{Conclusion}
\label{sec:conclusion}

We introduced AdaFusion, a lightweight adaptive fusion framework for integrating knowledge from multiple, diverse PFMs. By leveraging a trainable gating module over low-dimensional, frozen features, AdaFusion consistently outperforms individual PFMs and established fusion baselines across classification and regression benchmarks. Crucially, AdaFusion offers contribution-driven interpretability that provides evidence consistent with model-specific preferences across tissue phenotypes, offering a practical mechanism to both improve performance and better characterise model-specific inductive biases. Future directions include automating PFM ensemble selection and further linking contribution patterns to fine-grained histological structures.

\begin{credits}
\subsubsection{\ackname} This work was supported by the following grants. JR is supported by the NIHR Oxford Biomedical Research Centre. KY is supported by the National Natural Science Foundation of China (No. 62476101) and the Guangdong Basic and Applied Basic Research Foundation (Grant No. 2024A1515140137). 
The views expressed are those of the authors and not necessarily those of the NHS, the NIHR, or the Department of Health.

\subsubsection{\discintname}
The authors have no competing interests to declare that are relevant to the content of this article.
\end{credits}
\newpage

\bibliographystyle{splncs04}
\bibliography{paper-0003}

\end{document}